\documentclass[preprint,12pt]{elsarticle}

\usepackage{amssymb}
\usepackage{amsmath,amsfonts}
\usepackage{booktabs}
\usepackage{array}
\usepackage{algpseudocode}
\usepackage{graphicx}
\usepackage{float}
\usepackage{textcomp}
\usepackage{stfloats}
\usepackage{url}
\usepackage{verbatim}
\usepackage{subcaption}
\usepackage{xcolor}
\usepackage{multirow}
\usepackage[T1]{fontenc}
\usepackage[justification=centering]{caption}
\usepackage{makecell}

\journal{Neurocomputing}

\begin{document}

\begin{frontmatter}

\title{A Geometric Probe of the Accuracy-Robustness Trade-off: Sharp Boundaries in Symmetry-Breaking Dimensional Expansion}

\tnotetext[fund]{This work was supported by the National Natural Science Foundation of China (Grant No. 72501224).}

\author[inst1]{Yu Bai}
\author[inst1]{Zhe Wang}
\author[inst1]{Jiarui Zhang}
\author[inst1]{Dong-Xiao Zhang}
\author[inst1]{Yinjun Gao\corref{cor1}}
\ead{gaoyinjun@nint.ac.cn}
\author[inst1]{Jun-Jie Zhang\corref{cor1}}
\ead{zjacob@mail.ustc.edu.cn}

\cortext[cor1]{Corresponding authors.}

\affiliation[inst1]{organization={Northwest Institute of Nuclear Technology},
            city={Xi'an},
            postcode={710024},
            country={P.R. China}}

\begin{abstract}
The trade-off between clean accuracy and adversarial robustness is a pervasive phenomenon in deep learning, yet its geometric origin remains elusive. In this work, we utilize Symmetry-Breaking Dimensional Expansion (SBDE) as a controlled probe to investigate the mechanism underlying this trade-off. SBDE expands input images by inserting constant-valued pixels, which breaks translational symmetry and consistently improves clean accuracy (e.g., from $90.47\%$ to $95.63\%$ on CIFAR-10 with ResNet-18) by reducing parameter degeneracy. However, this accuracy gain comes at the cost of reduced robustness against iterative white-box attacks. By employing a test-time \emph{mask projection} that resets the inserted auxiliary pixels to their training values, we demonstrate that the vulnerability stems almost entirely from the inserted dimensions. The projection effectively neutralizes the attacks and restores robustness, revealing that the model achieves high accuracy by creating \emph{sharp boundaries} (steep loss gradients) specifically along the auxiliary axes. Our findings provide a concrete geometric explanation for the accuracy-robustness paradox: the optimization landscape deepens the basin of attraction to improve accuracy but inevitably erects steep walls along the auxiliary degrees of freedom, creating a fragile sensitivity to off-manifold perturbations.
\end{abstract}


\end{frontmatter}

\section{Introduction}

Deep neural networks (DNNs) have established themselves as the standard tools in computer vision and beyond, delivering remarkable performance in both benchmarks and real-world deployments~\cite{lecun2015deep, he2016deep, dosovitskiy2021an, krizhevsky2017imagenet, simonyan2015very, Sun_2024}. However, this success is shadowed by a persistent vulnerability: small, imperceptible input perturbations crafted by an adversary can flip model predictions with high confidence~\cite{szegedy2014intriguing, goodfellow2015explaining, carlini2017towards, papernot2016limitations}. This phenomenon—exceptional clean accuracy coexisting with extreme fragility—has motivated extensive research into the geometry of decision boundaries.

A central theme in this literature is the apparent trade-off between clean accuracy and adversarial robustness. Theoretical and empirical studies suggest that these two goals may be inherently conflicting in high-dimensional spaces, or at least require significantly more data to reconcile~\cite{tsipras2019robustness, zhang2019theoretically, schmidt2018adversarially}. Early geometric analyses showed that decision boundaries often lie dangerously close to natural samples~\cite{szegedy2014intriguing, carlini2017towards, fawzi2018cvpr, gilmer2018spheres, stutz2019cvpr}, while others pointed to local linearity and gradient accumulation as culprits~\cite{goodfellow2015explaining, akhtar2018threat}. More recent perspectives argue that models maximize accuracy by latching onto "non-robust" features that are predictive but brittle~\cite{ilyas2019adversarial}. Furthermore, the evaluation of robustness itself is fraught with difficulty, as stronger attacks often negate apparent gains from proposed defenses~\cite{athalye2018obfuscated, tramer2020adaptive, meng2022Adversarial, wang2022Adversarial}. Despite these insights, it is still beneficial to provide a clear, controllable geometric picture of \emph{how} the loss landscape deforms as a model is pushed toward higher accuracy, and exactly \emph{why} this deformation leads to fragility.

In this work, we investigate this trade-off by using a specific input transformation —Symmetry-Breaking Dimensional Expansion (SBDE)— as a \emph{controlled probe}. SBDE expands the input image by inserting constant-valued pixels at fixed intervals, a procedure that breaks translational symmetry and aids optimization by reducing parameter degeneracy~\cite{symmetry2025}. We use SBDE as an experimental "knob" to systematically raise clean accuracy (e.g., from $90.47\%$ to $95.63\%$ on CIFAR-10 with ResNet-18) while keeping the training pipeline unchanged. This setup allows us to isolate the geometric consequences of improving accuracy.

Our experiments reveal a striking geometric phenomenon. While SBDE successfully creates a deeper basin of attraction for natural data (improving clean accuracy), it simultaneously erects \textbf{sharp boundaries}—extremely steep loss gradients—along the newly introduced auxiliary dimensions. When we subject the model to iterative white-box attacks, we observe that the adversarial perturbations concentrate almost entirely on these auxiliary coordinates, exploiting the steep slopes to maximize loss. 

To verify that these sharp boundaries are indeed the source of the trade-off, we introduce a test-time \textbf{mask projection} that resets the auxiliary pixels to their training constants. This simple operation neutralizes the attack and restores robustness to a high level (see Table~\ref{tab:accuracy_tradeoff_cifar_10}), confirming that the vulnerability was localized to the auxiliary subspace. These findings support a concrete geometric explanation for the accuracy-robustness paradox: the pursuit of higher accuracy drives the model to carve out a narrower, more anisotropic loss landscape, where stability on the signal manifold is purchased at the cost of steep, fragile walls in the auxiliary directions. This interpretation aligns with recent uncertainty-based perspectives on the tension between accuracy and robustness~\cite{zhang2024exploring, ZHANG2025112197, zhang2024nsr}.

\section{Method: a controlled probe of loss geometry}
\label{sec:method}

To dissect the geometric mechanisms underlying the accuracy-robustness trade-off, we treat the training and evaluation process as a controlled physical experiment rather than a standard benchmark competition. Our methodology consists of three integrated components, forming a "Construction-Probe-Verification" pipeline:

\begin{enumerate}
    \item \textbf{Construction (SBDE):} We first employ SBDE to modify the input space. This acts as our experimental instrument to systematically enhance clean accuracy and reshape the loss landscape.
    \item \textbf{Probe (Iterative Attacks):} Instead of using attacks merely to break the model, we utilize iterative gradient-based methods (e.g., PGD) as probes to trace the directions of steepest ascent in this reshaped landscape.
    \item \textbf{Verification (Mask Projection):} We introduce a test-time projection operator $\Pi$ to decouple the signal dimensions from the auxiliary ones. This allows us to rigorously verify whether the observed vulnerability stems from the signal manifold or the artificially introduced auxiliary boundaries.
\end{enumerate}

The mathematical formulation and geometric intuition for each component are detailed below.

\subsection{Symmetry-breaking expansion: motivation and operator}
\label{subsection: motivation}
\textbf{Symmetry breaking as an optimization aid.}
Symmetry breaking is a fundamental concept in physics: perturbing a symmetric system can smooth the budget landscape and reduce degenerate minima. In neural networks, augmenting the input with constant-valued dimensions can mimic such a process and facilitate more stable gradient-based optimization~\cite{symmetry2025}. Consider a two-layer network with input \(x\), weights \(w,w'\), and activation \(\sigma\):
\begin{eqnarray}
\sigma(\sigma(w_{i,j} x_{i})\,w^{\prime}_{j,1})
&\sim& w_{i,j} x_{i} w^{\prime}_{j,1}
+\mathcal{O}(\text{higher orders}), \label{eq:output_layers}
\end{eqnarray}
where repeated indices are summed. The mapping is symmetric under permutations \(w_{i,j_\alpha}\!\leftrightarrow\! w_{i,j_\beta}\) and \(w'_{j_\alpha,1}\!\leftrightarrow\! w'_{j_\beta,1}\), which yields many equivalent minima. Adding a constant-valued input coordinate \(x_{\mathrm c}\) introduces new parameters \(w^{\prime\prime}\),
\begin{eqnarray}
\sigma(\sigma(w_{i,j} x_{i}+w^{\prime\prime}_{1,j} x_{\mathrm c})\,w^{\prime}_{j,1})
&\sim& w_{i,j} x_{i} w^{\prime}_{j,1}
+ w^{\prime\prime}_{1,j} x_{\mathrm c} w^{\prime}_{j,1}
+\mathcal{O}(\text{higher orders}), \label{eq:output_2_layers}
\end{eqnarray}
which breaks the permutation symmetry and reduces degeneracy, akin to the role of external fields in the Ising model~\cite{Ising1925}.

\textbf{SBDE operator on images.}
For image classification, let \(x\!\in\!\mathbb{R}^{C\times H\times W}\) and \(f_\theta\) a classifier. SBDE deterministically maps \(x\) to an expanded tensor \(\tilde{x}\!\in\!\mathbb{R}^{C\times (HF)\times (WF)}\) by interleaving constant-valued rows/columns with spacing \(F\in\{2,3,\dots\}\). The expanded grid naturally partitions coordinates into two disjoint sets:
\[
\Omega_{\text{sig}}=\text{(embedded original pixels)},\qquad
\Omega_{\text{aux}}=\text{(inserted constant pixels)}.
\]

\begin{figure}[t]
\centering
\begin{minipage}{1\textwidth}
    \centering
    \includegraphics[width=0.7\textwidth]{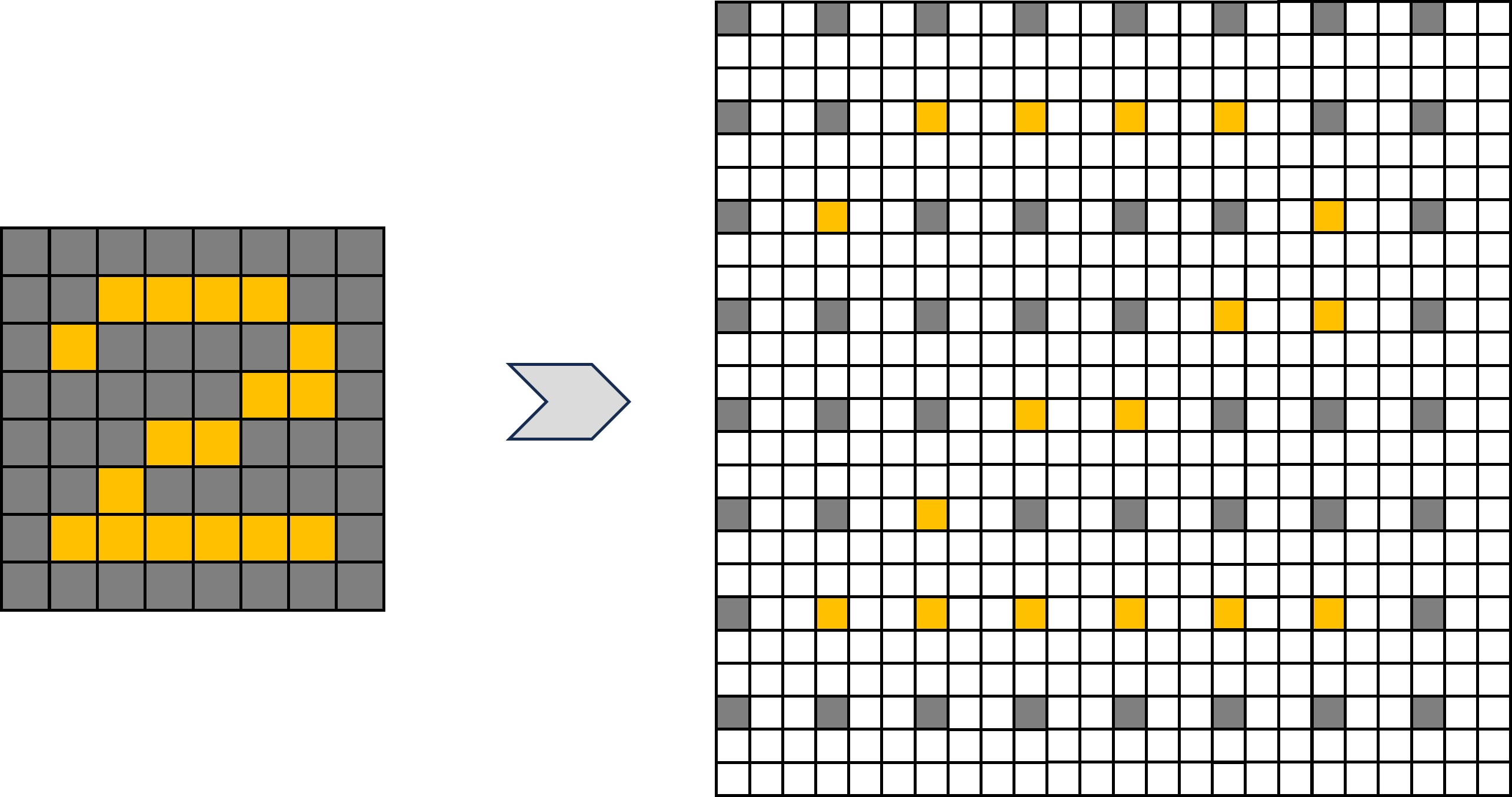}
    \caption{Symmetry-breaking dimensional expansion (illustration). Original image pixels are mapped to designated locations in the expanded grid; constant-valued rows/columns fill the rest.}
    \label{fig:non_centered_expansion}
\end{minipage}
\end{figure}

\begin{figure}[H]
    \centering
    \includegraphics[width=0.8\linewidth]{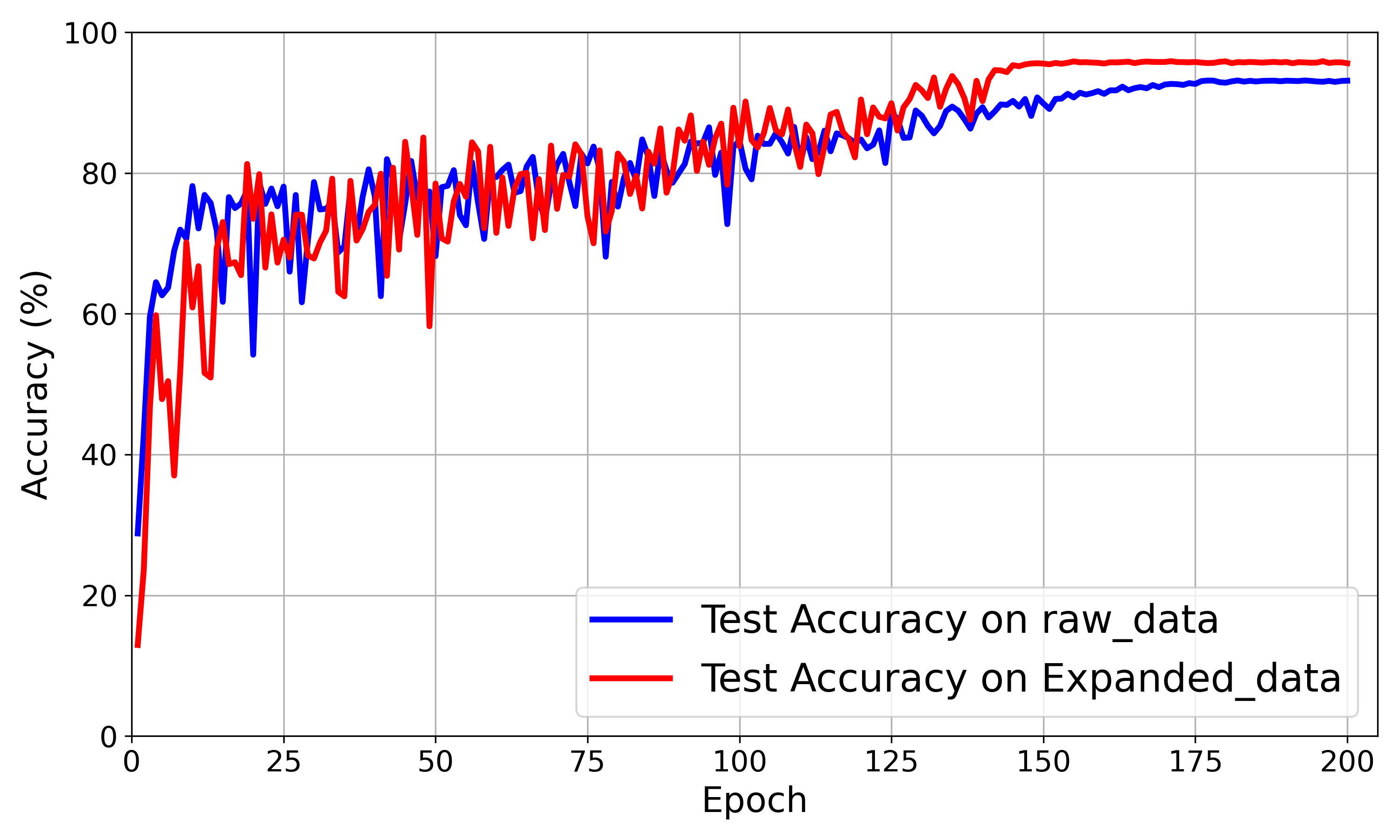}
    \caption{Test accuracy trajectory on raw vs.\ SBDE-expanded data (ResNet-18 on CIFAR-10).}
    \label{fig:accurate_SBDE}
\end{figure}

During training we \emph{always} keep \(\tilde{x}(i,j)=c\) for \((i,j)\!\in\!\Omega_{\text{aux}}\), where \(c\) is a fixed constant (e.g., \(0\) or dataset mean). The typical placement is illustrated in Figure~\ref{fig:non_centered_expansion}. Empirically, SBDE consistently improves clean accuracy across the training trajectory (Figure~\ref{fig:accurate_SBDE}), aligning with~\cite{symmetry2025}. 

Crucially, from a geometric standpoint, while SBDE reduces parameter degeneracy and helps the optimization find a deeper minimum (higher accuracy), it implicitly imposes a strict constraint: the model learns to expect constant values at $\Omega_{\text{aux}}$. This strong constraint suggests the formation of a "narrower" basin, where deviations along $\Omega_{\text{aux}}$ might encounter steep loss walls—a hypothesis we test in the next section.

\subsection{Iterative White-Box Attacks as a Local Geometry Probe}

Iterative white-box attacks, such as the \(\ell_\infty\)-bounded Projected Gradient Descent (PGD) \cite{madry2018towards} attack, offer a powerful way to probe the local geometry of the loss landscape. Given a trained classifier \( f_\theta \) and a loss function \( L(\cdot, y) \), we apply PGD in the expanded space (i.e., the space where SBDE has been applied to the input images).

The iterative attack is conducted as follows:
\begin{equation}
\tilde{x}^{(t+1)} \leftarrow 
\Pi_{\mathcal{B}_\infty(\tilde{x}, \varepsilon)} \left( 
    \tilde{x}^{(t)} + \alpha \, \mathrm{sign}\left( \nabla_{\tilde{x}} L(f_\theta(\tilde{x}^{(t)}), y) \right)
\right),
\label{eq:pgd}
\end{equation}
where \( \tilde{x}^{(0)} = \tilde{x} \) is the initial image in the expanded space, and \( \alpha \) is the step size. The gradient \( \nabla_{\tilde{x}} L \) computes the direction in which the loss increases most rapidly with respect to the input image. The sign of the gradient determines the direction of perturbation to the input image at each iteration. The perturbation is constrained to the \( \ell_\infty \)-ball, ensuring the attack remains within a perturbation budget \( \varepsilon \). We also evaluate Auto Projected Gradient Descent (APGD) \cite{croce2020reliable}, Auto Projected Gradient Descent – Targeted (APGDT) \cite{croce2020reliable}, Basic Iterative Method (BIM) \cite{Kurakin2018Adversarial} and AutoAttack \cite{croce2020reliable}.

Instead of viewing iterative attacks merely as security threats, we treat them as \emph{gradient tracers} that map the steepest ascent directions of the loss landscape. In the expanded space, the gradient $\nabla_{\tilde{x}} L$ has components along both signal directions ($\Omega_{\text{sig}}$) and auxiliary directions ($\Omega_{\text{aux}}$).

Standard training forces $\Omega_{\text{aux}}$ to be constant, but it puts no penalty on the \emph{gradients} with respect to these coordinates. Consequently, while the loss surface is flat at the training point, it may curve upward explicitly sharply as soon as one moves away along the auxiliary axes. We hypothesize that iterative attacks, which greedily follow gradients, will be "magnetized" by these steep slopes, concentrating their perturbation budget $\varepsilon$ almost entirely on the auxiliary dimensions.

\subsection{Mask Projection for Verification}

To verify the findings from the gradient-based attack and better understand the influence of the SBDE-modified input, we introduce a mask projection \( \Pi \). This simple yet effective operation allows us to isolate and evaluate the effect of the auxiliary pixels (those added by SBDE) on the model’s performance.

The projection operation is defined as follows:
\begin{equation}
\Pi(x)(i,j) =
\begin{cases}
x(i,j), & (i,j) \in \Omega_{\text{sig}}, \\
c, & (i,j) \in \Omega_{\text{aux}},
\end{cases}
\label{eq:projection}
\end{equation}
where $x(i,j)$ denotes the pixel value of the input image $x$ at coordinates $(i,j)$, and $c$ is a constant padding value. Here, $\Omega_{\text{sig}}$ represents the set of signal pixels (i.e., the original pixels unaltered by the SBDE expansion), while $\Omega_{\text{aux}}$ denotes the set of auxiliary pixels inserted during the SBDE process.

The mask projection \( \Pi \) resets all auxiliary pixels to their constant value \( c \), which is the same value they had during training, while keeping the signal pixels unchanged. This operation effectively restores the auxiliary coordinates to their training condition, allowing us to observe whether perturbations concentrated on these auxiliary coordinates were responsible for the loss in robustness.

Geometrically, $\Pi$ acts as a projection operator that snaps the perturbed point back onto the hyperplane defined by $x(i,j)=c, \forall (i,j) \in \Omega_{\text{aux}}$. If the SBDE-induced vulnerability is indeed caused by steep "loss walls" erected along the auxiliary dimensions, then the adversarial example $\tilde{x}_{adv}$ should lie high up on these walls. Applying $\Pi$ removes the displacement along these steep directions, effectively sliding the point back down into the flat basin of the signal manifold. A successful recovery of accuracy by $\Pi$ thus serves as definitive proof that the loss landscape is highly anisotropic: steep along $\Omega_{\text{aux}}$, but robust along $\Omega_{\text{sig}}$.

This simple procedure enables us to isolate the impact of the auxiliary coordinates on adversarial vulnerability. By comparing the model's performance before and after applying the projection \( \Pi \), we can evaluate whether the adversarial effects were concentrated on the added auxiliary dimensions. A significant recovery in robustness after applying the projection supports the hypothesis that the auxiliary coordinates played a key role in the observed loss of robustness.

\subsection{Experimental Setup}
\label{sec:setup}

We follow standard practices for robustness evaluation as outlined in prior works~\cite{madry2018towards, croce2020reliable, tramer2020adaptive}. In this study, we aim to compare the performance of \textbf{SBDE} and \textbf{SBDE + \(\Pi\)} defenses and analyze the trade-off between the improvements in accuracy achieved by SBDE and the potential reduction in robustness. From a geometric perspective, we hypothesize that the loss function becomes steeper at the expanded pixel locations, which impacts the model's robustness. To investigate these points, we present tables with the robust accuracy of SBDE across seven attack methods, as well as the comparison of clean accuracy and adversarial robustness under SBDE.

\paragraph{\textbf{Model and Data}}
For our experiments, we use ResNet-18, modified to accommodate expanded-pixel inputs. The first convolution layer is a \(7 \times 7\) kernel with stride 2, and the initial max-pool layer is replaced by an identity function, allowing the network to handle the larger spatial sizes introduced by pixel expansion. The model's final fully connected layer is re-initialized for the 10 classes of the CIFAR-10 dataset.

We explore six insertion schemes, including constant values of 0.0, 0.5, 0.2, and 0.3, as well as two cyclic insertion patterns, 0.2-GapCycle and 0.3-GapCycle. The 0.2-GapCycle pattern alternates values of -0.2, 0, and 0.2 in a cyclic manner at each insertion location, creating an alternating fill pattern across the expanded space. The same alternating scheme applies to 0.3-GapCycle.

\paragraph{\textbf{Training}}
Training is carried out using stochastic gradient descent (SGD) with momentum set to 0.9 and weight decay of \(5 \times 10^{-4}\). A cosine-annealing learning rate schedule is employed, starting at 0.1 and decaying to \(1 \times 10^{-5}\) over 200 epochs. The batch size is set to 256, and mixed-precision training is used to accelerate the process via automatic mixed precision with \texttt{autocast} and gradient scaling with \texttt{GradScaler}. Random seeds are fixed to ensure reproducibility.

For data augmentation, we apply random cropping with a 4-pixel padding and random horizontal flipping, consistent with standard practices for CIFAR-10. The model is trained from random initialization for the full 200-epoch schedule.

\paragraph{\textbf{Attacks}}
We evaluate the robustness of our models using a range of adversarial attacks. All attacks are conducted in the \textbf{expanded} input space. Unless specified otherwise, we use \(\ell_\infty\)-bounded attacks with \(\varepsilon = 8/255\). The following attacks are employed:

- \textbf{PGD} and \textbf{APGD}: Both attacks are executed with 20 steps.

- \textbf{BIM}: This attack uses 10 steps.

- \textbf{AutoAttack}: We also evaluate the model against the comprehensive AutoAttack suite~\cite{croce2020reliable}.

- Additional attacks, such as \textbf{APGDT}, are included in our evaluation.

For the \textbf{SBDE} evaluations, attacks are applied directly in the expanded space. In contrast, for the \textbf{SBDE + \(\Pi\)} evaluations, we apply the projection operator \(\Pi\) (Eq.~\eqref{eq:projection}) before inference, effectively applying an additional mask after the attack in the expanded space. 

\paragraph{\textbf{Threat Model}}
Our diagnostic framework assumes that the attacker perturbs the model in the expanded space and does not differentiate through \(\Pi\). This approach follows standard preprocessing-style evaluations. The primary goal is to perform a geometric analysis of the effects of pixel expansion on model robustness, rather than claim an undefeatable defense.

\section{Results}
\label{sec:results}

This section explains \emph{what} SBDE changes in practice and \emph{why} those changes occur, using the measurements in Table~\ref{tab:accuracy_tradeoff_cifar_10} and the perturbation visualization in Figure~\ref{fig:pgd_mask}. We proceed from phenomena to mechanism: we first document the accuracy/robustness outcomes, then identify where attacks act in the expanded space, and finally connect the behavior to a concrete geometric account of the loss landscape under SBDE. Ablations on the fill constant and expansion factor corroborate that this account is structural rather than brittle.

\subsection{SBDE raises clean accuracy yet reduces robustness}

Training with SBDE reliably improves clean accuracy while leaving the rest of the pipeline unchanged. On CIFAR-10 with ResNet-18, clean accuracy increases from \(90.47\%\) (Natural) to \(95.63\%\) (SBDE), matching the training trajectory trend in Figure~\ref{fig:accurate_SBDE} and reproducing prior reports of accuracy gains with symmetry breaking. Intuitively, SBDE inserts constant-valued rows/columns at fixed intervals, which breaks symmetry of the parameter space and makes the convolutional stem more position-aware. This reduces parameter degeneracy (many weight configurations no longer represent the same loss function value), improves gradient informativeness, and helps optimization settle into higher-accuracy minima. 

When we test robustness in the \emph{expanded} input space, we first observe a clear empirical fact: the SBDE-trained model, though achieving higher clean accuracy, suffers a drop in robust accuracy under strong iterative white-box attacks (Table~\ref{tab:accuracy_tradeoff_cifar_10}, row "SBDE (without \(\Pi\))"). In other words, small adversarial perturbations now cause larger changes in the model’s prediction. 

To understand why, it is helpful to picture the loss surface as a landscape over the input space. Training with SBDE effectively reshapes this landscape. The added constant pixels act like anchors that guide the network toward more consistent, high-accuracy solutions—similar to forming a deeper "valley" around each correctly classified input. However, this deeper valley also becomes narrower: the loss increases much faster when moving slightly away from the data point in certain directions. These steep directions correspond to the new auxiliary (inserted) pixels introduced by SBDE. 

Iterative attacks such as PGD or AutoAttack search for exactly these directions of fastest increase in the loss. Because the SBDE landscape has sharper slopes along the auxiliary coordinates, the attacks quickly climb those slopes, causing misclassification even when the visible image content changes only imperceptibly. Thus, the same geometric reshaping that improves clean accuracy (by creating a deeper, more focused basin around the data) also makes the model more sensitive to small, adversarial perturbations in specific directions. This observation is consistent with the "uncertainty principle of neural networks" \cite{zhang2024exploring, ZHANG2025112197, zhang2024nsr, Zhang_2025}. 

\begin{table}[H]
\centering
\caption{CIFAR-10, ResNet-18. Highly accurate neural network is more fragile under attacks. Using projection $\Pi$ at inference strongly mitigates iterative attacks on SBDE models.}
\label{tab:accuracy_tradeoff_cifar_10}
\resizebox{\textwidth}{!}{ 
\begin{tabular}{lccccccc}
\toprule
\textbf{Method} & \textbf{Clean}  & \textbf{PGD} & \textbf{AutoAttack} & \textbf{BIM} & \textbf{APGD}  & \textbf{APGDT} & \textbf{AVG} \\
\midrule
Natural    & 90.47  & 0.07   & 0.01   & 0.11   & 0.10      & 0.02 & 0.06 \\
SBDE (without $\Pi$)   & 95.63   & 0.00  & 0.00 & 0.00 & 0.00  & 0.00 & 0.00 \\
\textbf{SBDE (with $\Pi$)}   & \textbf{95.63}  & \textbf{88.11} & \textbf{86.25} & \textbf{60.02} & \textbf{88.10}  & \textbf{86.86} & \textbf{81.87} \\
\bottomrule
\end{tabular}
}
\end{table}

\subsection{Projection \(\Pi\) cancels the attack while preserving accuracy}

Table~\ref{tab:accuracy_tradeoff_cifar_10} (row "SBDE (with \(\Pi\))") shows a striking recovery: after applying the simple projection \(\Pi\), the SBDE model retains its high clean accuracy (\(95.63\%\)) and regains strong robustness across all iterative attacks (average \(\approx 85\%\)). The only change introduced by \(\Pi\) is to restore each auxiliary coordinate in \(\Omega_{\text{aux}}\) to its original training constant \(c\) \emph{right before inference}, leaving the genuine signal coordinates \(\Omega_{\text{sig}}\) untouched. Note that the recovered robust accuracy is slightly lower than the clean accuracy. This indicates that while the dominant failure mode is driven by the auxiliary dimensions, a small fraction of the attack budget does perturb the signal features, albeit with much lower effectiveness.

This observation provides a powerful clue about the geometry of the loss surface. If the projection almost entirely cancels the attack, it means that most adversarial perturbations were confined to the auxiliary subspace. In other words, the attack had been "climbing" steep slopes that exist primarily along those synthetic coordinates introduced by SBDE.

\begin{figure}[H]
    \centering
    \includegraphics[width=\linewidth]{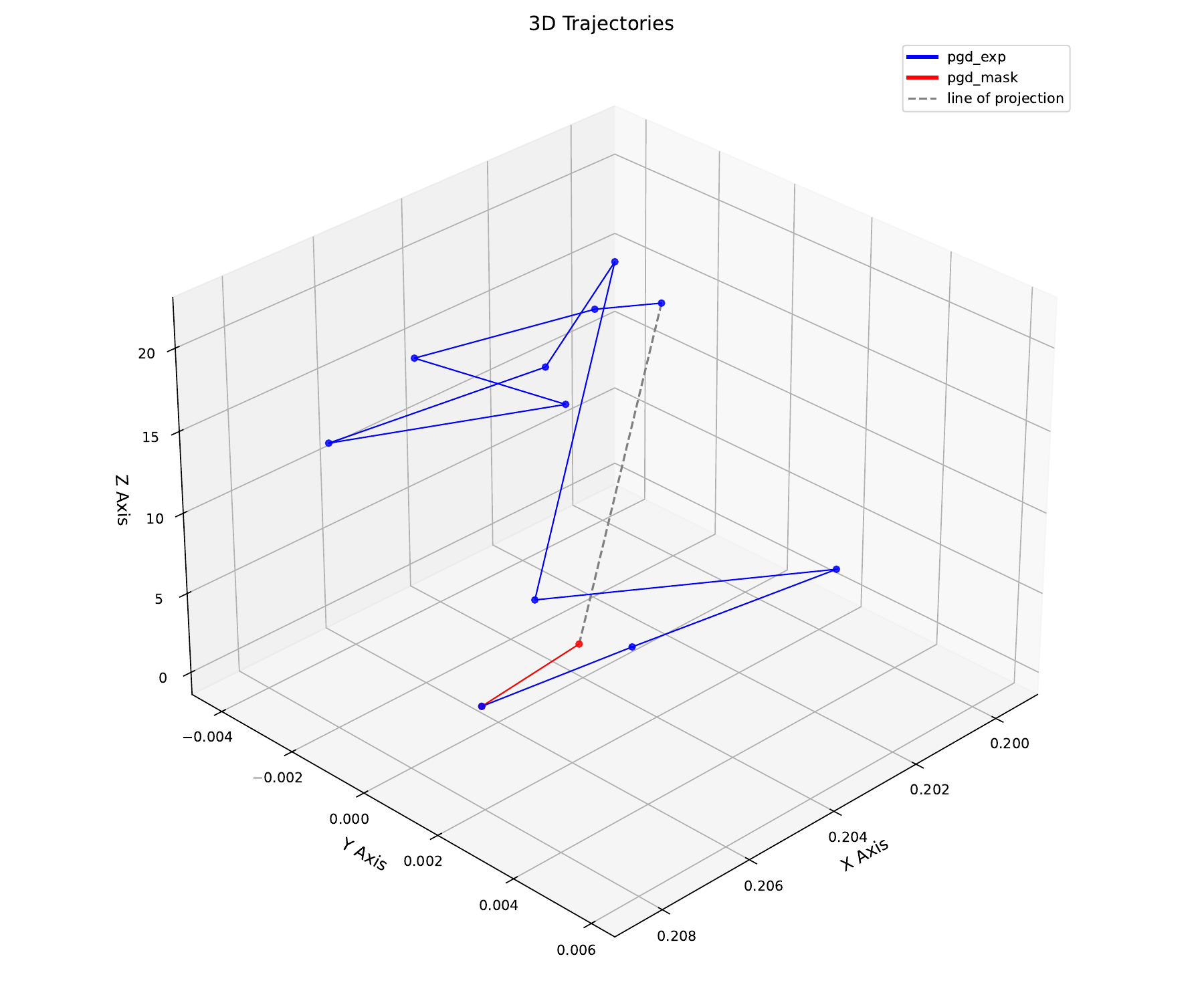}
    \caption{Perturbation trajectory after an iterative attack. The Z-axis corresponds to the value of the loss function for the attacked image at each iterative step. The X-axis and Y-axis correspond to the pixel value of a randomly selected pixel and its adjacent auxiliary pixel, respectively. Blue lines represent the iterative attacks of PGD (with the increase of the step, the loss value increases). Red line links the start point (no attack) and end point (Projection of attacked image).  After projection, the loss function drops prominently, indicating that high-magnitude pixels concentrate on SBDE’s auxiliary coordinates $\Omega_{\text{aux}}$.}
    \label{fig:pgd_mask}
\end{figure}

To make this phenomenon visible, Figure~\ref{fig:pgd_mask} visualizes how iterative attacks navigate the expanded input space. On the horizontal axis (\(x\)) we pick a pixel from the original signal region; on the vertical axis (\(y\)) we place its neighboring auxiliary pixel inserted by SBDE; and on the vertical axis (\(z\)) we plot the loss value at each step of the attack. Each blue line shows how a standard PGD iteration moves in this \((x,y)\) plane. We can see that the blue trajectories climb steeply upward.

When we apply the same PGD attack but then enforce \(\Pi\) \emph{after} the attack—i.e., we take the final adversarial example and reset all auxiliary coordinates to the constant value \(c\)—the picture changes dramatically. In Figure~\ref{fig:pgd_mask}, this operation is represented by a single red line that jumps from the final adversarial point back onto the plane \(y=c\). This line shows how the mask projection collapses the auxiliary dimensions: all the steep loss contributions that the PGD attack had accumulated along the \(y\) (auxiliary) axis are suddenly removed. The endpoint of the red line lies at a much lower loss value than the end of the blue PGD trajectory, illustrating that once the auxiliary coordinates are restored, the adversarial perturbation almost completely loses its effect.

Geometrically, this means that the projection \(\Pi\) erases the steep "ridges" of the loss surface that run along the auxiliary directions \(\Omega_{\text{aux}}\). What remains is the smoother basin formed by the signal coordinates \(\Omega_{\text{sig}}\), whose slopes correspond to genuine semantic variations in the input image. Because iterative attacks like PGD and APGD are local gradient-following procedures, they had exploited precisely those steep ridges to increase the loss. When \(\Pi\) flattens them, the attack loses its most effective ascent path, and the model’s prediction returns to its original, correct class.

This visual evidence makes the underlying mechanism explicit. The degradation of robustness in the unprojected SBDE model does not stem from fragile features in the original signal; instead, it arises from artificially steep gradients concentrated in the auxiliary subspace that SBDE introduced to break symmetry. By reinstating the training constraint through \(\Pi\), we suppress those steep directions and reveal that the classifier remains stable within the true signal manifold.

In this light, the projection experiment turns SBDE into a \emph{diagnostic probe} of the loss landscape. By expanding the input space, letting attacks explore it, and then projecting back onto the trained manifold, we can directly observe where the network’s loss is steep and how that steepness relates to both accuracy and robustness. SBDE thus provides a controlled geometric lens: it shows that the gain in clean accuracy comes from sharper curvature along constructed auxiliary dimensions, while robustness drops only because iterative attacks exploit those very directions.

\subsection{Ablations}

To further verify that the behaviors we observe are not accidental, we conducted extensive ablations on two dimensions of the SBDE design: (i) the value of the inserted constant \(c\) and its filling pattern, (ii) the interaction between the expansion factor \(F\) and the sampling geometry of the first convolutional stem (with stride settings of 2, 3, and 4). These ablations allow us to test whether the observed accuracy–robustness trade-off and the concentration of attack budget on auxiliary coordinates persist across diverse configurations.

\vspace{0.3em}
\noindent
\textbf{Effect of the filling constant and expansion placement.}
To test whether the observed SBDE behavior depends on the specific way auxiliary pixels are filled or placed, we vary the constant value \(c\) inserted into the auxiliary coordinates—choosing among \(0.0, 0.2, 0.3,\) and \(0.5\)—and also introduce simple periodic “gap-cycling’’ patterns (such as alternating constants \(0.2\)-cycle or \(0.3\)-cycle). The corresponding results are reported in Table~\ref{tab:Non-Centered_expansion_robust_accuracy_cifar_10_constant_value}. 

\begin{table}[H]
\centering
\caption{SBDE: varying the fill constant $c$.}
\label{tab:Non-Centered_expansion_robust_accuracy_cifar_10_constant_value}
\resizebox{\textwidth}{!}{ 
\begin{tabular}{lccccccc}
\toprule
\textbf{Fill Constant} & \textbf{Clean}  & \textbf{PGD} & \textbf{AutoAttack} & \textbf{BIM} & \textbf{APGD}  & \textbf{APGDT} & \textbf{AVG} \\
\midrule
0.0 (without $\Pi$)  & 95.63   & 0.00  & 0.00 & 0.00 & 0.00 & 0.00 & 0.00 \\
\textbf{0.0 (with $\Pi$)}   & \textbf{95.63}  & \textbf{88.11}   & \textbf{86.25}   & \textbf{60.02}   & \textbf{88.10}   & \textbf{86.86} & \textbf{85.88} \\
0.5 (without $\Pi$)  & 95.65  & 0.00  & 0.00 & 0.00 & 0.00  & 0.00 & 0.00 \\
\textbf{0.5 (with $\Pi$)}     & \textbf{95.65}   & \textbf{86.49}  & \textbf{80.62}  & \textbf{56.85}  & \textbf{83.60}  & \textbf{83.96} & \textbf{83.26} \\
0.2 (without $\Pi$)  & 95.34   & 0.00 & 0.00 & 0.00 & 0.00  & 0.00 & 0.00 \\
\textbf{0.2 (with $\Pi$)}     & \textbf{95.34}   & \textbf{86.59}  & \textbf{85.98}  & \textbf{56.43}  & \textbf{88.15}  & \textbf{79.22} & \textbf{83.86} \\
0.3 (without $\Pi$)  & 95.44  & 0.00 & 0.00 & 0.00 & 0.00  & 0.00 & 0.00 \\
\textbf{0.3 (with $\Pi$)}     & \textbf{95.44}   & \textbf{85.73}  & \textbf{84.68}  & \textbf{56.11}  & \textbf{86.63}  & \textbf{84.65} & \textbf{84.08} \\
0.2-GapCycle (without $\Pi$) & 95.45  & 0.00 & 0.00 & 0.00 & 0.00  & 0.00 & 0.00 \\
\textbf{0.2-GapCycle (with $\Pi$)}   & \textbf{95.45}   & \textbf{86.28}  & \textbf{84.26}  & \textbf{58.20}  & \textbf{87.50}  & \textbf{84.49} & \textbf{84.57} \\
0.3-GapCycle (without $\Pi$) & 95.40  & 0.00 & 0.00 & 0.00 & 0.00 & 0.00 & 0.00 \\
\textbf{0.3-GapCycle (with $\Pi$)}   & \textbf{95.40}   & \textbf{86.83}  & \textbf{85.03}  & \textbf{55.48}  & \textbf{87.43}  & \textbf{86.98} & \textbf{84.70} \\
\bottomrule
\end{tabular}
}
\end{table}

Across all these variants, the same pattern consistently emerges. Clean accuracy remains high—typically above \(95\%\)—and projected robust accuracy after applying \(\Pi\) is nearly unchanged across different choices of \(c\) and placements. Meanwhile, robustness before projection always drops sharply under iterative attacks, confirming that the attacks predominantly exploit the auxiliary coordinates. In other words, as long as the inserted constant lies within the normal pixel range of the training data, SBDE reliably exhibits the same “accuracy–robustness divergence’’: accuracy improves, robustness declines under direct attack, and robustness is largely restored once we project the auxiliary coordinates back to their fixed value. 

This stability across both constant values and spatial placements indicates that the geometric mechanism of SBDE does not hinge on any special numeric choice or particular layout. What matters is the \emph{existence of a fixed, predictable lattice} of auxiliary coordinates that remain constant during training. The network internalizes this invariance and encodes it into its representation, effectively learning that deviations along those auxiliary directions correspond to large increases in loss. It is this learned constraint—rather than the specific fill number or padding pattern—that sculpts the anisotropic loss landscape where gradients are steep along the auxiliary dimensions and relatively flat along the true signal manifold.

\vspace{0.3em}
\noindent
\textbf{Expansion factor and network stride.}
We next study how the expansion factor \(F\) interacts with the sampling geometry of the first convolutional stem. We sweep \(F\in\{2,3,4,5,6,7,8,9\}\) for three stems that share a \(7\times7\) kernel but differ in stride: \(2\), \(3\), and \(4\). The results are reported in Tables~\ref{tab:Non-Centered_expansion_robust_accuracy_cifar_10_723}, \ref{tab:Non-Centered expansion robust_accuracy_cifar 10_733}, and \ref{tab:Non-Centered expansion robust_accuracy_cifar 10_743}, respectively. A clear and consistent regularity emerges: the post-projection robustness is noticeably \emph{worse} whenever \(F\) is an \emph{integer multiple} of the stem stride. Concretely, for stride \(=2\) the even factors \(F\in\{2,4,6,8\}\) underperform the odd factors \(F\in\{3,5,7,9\}\); for stride \(=3\) the multiples \(F\in\{3,6,9\}\) underperform the others; and for stride \(=4\) the multiples \(F\in\{2,4,8\}\) are weaker than the remaining choices.

\begin{table}[H]
\centering
\caption{SBDE, stride=2: expansion factor $F$ and alignment with the first convolution. Odd $F$ values align better in this setting.}
\label{tab:Non-Centered_expansion_robust_accuracy_cifar_10_723}
\resizebox{\textwidth}{!}{ 
\begin{tabular}{lccccccc}
\toprule
\textbf{\makecell{Expansion \\ Factor}} & \textbf{Clean}  & \textbf{PGD} & \textbf{AutoAttack} & \textbf{BIM} & \textbf{APGD}  & \textbf{APGDT} & \textbf{AVG} \\
\midrule
2    & 95.13  & 0.01    & 0.00    & 0.08    & 0.02     & 0.00  & 0.02 \\
3    & 95.55  & 81.57   & 83.37   & 56.47   & 86.80    & 72.50 & 76.14 \\
4    & 95.83  & 12.67   & 35.57   & 4.22    & 39.32    & 42.57 & 26.87 \\
5    & 95.63  & 88.11   & 86.25   & 60.02   & 88.10    & 86.86 & 81.87 \\
6    & 95.33  & 64.12   & 66.96   & 34.27   & 69.99    & 60.55 & 59.18 \\
7    & 95.03  & 80.63   & 80.42   & 47.81   & 79.18    & 86.51 & 74.91 \\
8    & 95.19  & 7.13    & 21.57   & 1.71    & 22.08    & 37.15 & 17.93 \\
9    & 94.52  & 82.91   & 70.30   & 33.87   & 73.57    & 84.68 & 69.07 \\
\bottomrule
\end{tabular}
}
\end{table}

This pattern admits a simple geometric explanation in terms of sampling alignment. The expanded grid produced by SBDE forms a periodic auxiliary lattice. When the stem stride divides \(F\), receptive fields land repeatedly on the \emph{same phase} of that lattice: the first-layer sampling “walks” the image in steps that resonate with the auxiliary period. This phase locking has two consequences. First, some transitions between signal and auxiliary coordinates are sampled less uniformly, which makes the training signal for the rule “auxiliary \(=\) constant” spatially uneven. Second, because early features see fewer cross-phase mixtures, the network internalizes the constraint less crisply: curvature along \(\Omega_{\text{aux}}\) is still larger than along \(\Omega_{\text{sig}}\), but the separation is blurred. In this regime iterative attacks leak a larger fraction of their budget into directions that are not purely auxiliary, and the projection \(\Pi\) removes a smaller share of the attack’s effect—hence the weaker post-projection robustness.

By contrast, when the stride and \(F\) are \emph{not} in an integer-multiple relation (e.g., odd \(F\) under stride \(2\)), the stem samples different phases of the auxiliary lattice more evenly. Kernels routinely span signal–auxiliary boundaries at varying offsets, making the “auxiliary \(=\) constant” constraint easy to learn everywhere. The resulting loss geometry is sharply anisotropic: gradients are steep along \(\Omega_{\text{aux}}\) and comparatively flat along \(\Omega_{\text{sig}}\). Iterative attacks then concentrate almost entirely on the auxiliary subspace, and \(\Pi\) cancels most of their effect, yielding the stronger post-projection robustness observed in the tables. In short, stride–expansion \emph{alignment} modulates the \emph{strength} of the phenomenon by controlling how crisply the auxiliary constraint is encoded, but it does not change its \emph{existence}: the accuracy gain from SBDE and the auxiliary-directed attack behavior persist across all configurations.

\begin{table}[t]
\centering
\caption{SBDE, stride=3: expansion factor $F$ and alignment with the first convolution. Odd $F$ values align better in this setting.}
\resizebox{\textwidth}{!}{ 
\begin{tabular}{lccccccc}
\toprule
\textbf{\makecell{Expansion \\ factor}} & \textbf{Clean}  & \textbf{PGD} & \textbf{AutoAttack} & \textbf{BIM} & \textbf{APGD}  & \textbf{APGDT} & \textbf{AVG} \\
\midrule
2    & 93.70  & 40.44   & 69.61   & 26.13   & 72.36   & 64.80 & 54.67 \\
3    & 95.10  & 0.00   & 0.00   & 0.07   & 0.01  & 0.00 & 0.02 \\
4    & 95.00  & 73.57   & 80.62   & 57.92   & 84.93  & 70.35 & 73.48 \\
5    & 95.70  & 89.51   & 88.40   & 61.89   & 89.27   & 86.87 & 83.19 \\
6    & 95.77  & 13.57   & 43.65   & 3.66   & 48.56   & 46.49 & 31.19 \\
7    & 95.47  & 90.39   & 86.73   & 62.22   & 86.80   & 89.94 & 83.22 \\
8    & 95.14  & 88.33   & 86.79   & 65.60   & 88.33   & 77.24 & 81.26 \\
9    & 95.47  & 57.36   & 69.36   & 30.65   & 72.28   & 61.79 & 58.29 \\
\bottomrule
\end{tabular}
}
\label{tab:Non-Centered expansion robust_accuracy_cifar 10_733}
\end{table}

\begin{table}[t]
\centering
\caption{SBDE, stride=4: expansion factor $F$ and alignment with the first convolution. Odd $F$ values align better in this setting.}
\resizebox{\textwidth}{!}{ 
\begin{tabular}{lccccccccc}
\toprule
\textbf{\makecell{Expansion \\ factor}} & \textbf{Clean}  & \textbf{PGD} & \textbf{AutoAttack} & \textbf{BIM} & \textbf{APGD}  & \textbf{APGDT} & \textbf{AVG} \\
\midrule
2    & 93.33 & 0.07  & 0.00  & 0.20  & 0.01   & 0.01  & 0.02 \\
3    & 94.18 & 79.24 & 84.02 & 60.36 & 86.13  & 76.07 & 77.16 \\
4    & 95.39 & 0.01  & 0.01  & 0.02  & 0.01   & 0.00  & 0.01 \\
5    & 95.44 & 92.12 & 91.37 & 69.32 & 92.25  & 90.54 & 87.12 \\
6    & 95.35 & 78.49 & 80.94 & 53.69 & 84.80  & 69.38 & 73.46 \\
7    & 95.40 & 92.85 & 91.02 & 71.61 & 91.66  & 92.41 & 87.91 \\
8    & 95.74 & 21.48 & 44.71 & 9.70  & 47.54  & 46.90 & 34.07 \\
9    & 95.23 & 90.93 & 90.48 & 68.69 & 91.08  & 91.38 & 86.51 \\
\bottomrule
\end{tabular}
}
\label{tab:Non-Centered expansion robust_accuracy_cifar 10_743}
\end{table}

\section{Discussion and Conclusion}

Accuracy and robustness have long been treated as two sides of a fragile balance. Empirical studies have cataloged this trade-off, and theoretical analyses have linked it to margin bounds or feature selection. Yet from a geometric perspective, we have lacked an intuitive, observable handle: a way to \emph{see} where accuracy gains bend the loss surface and how that bending creates vulnerability. The combination of SBDE and iterative attacks offers precisely such a handle.

SBDE acts as a controlled perturbation to the input geometry. By inserting constant-valued pixels, it explicitly defines a subspace---the auxiliary coordinates---where the loss curvature can develop independently of the natural image manifold. This manipulation turns an otherwise opaque learning process into a transparent one: as clean accuracy rises, we can literally watch the loss basin deepen and narrow, and we can measure where its steep walls appear. When iterative attacks like PGD trace their ascent through this expanded space, they reveal that the sharpest curvature resides along those synthetic auxiliary directions. Applying the projection \(\Pi\) then becomes a physical experiment: it slices away those directions and instantly removes the vulnerability. In this sense, SBDE converts the abstract idea of “loss anisotropy” into something measurable and visualizable.

This geometric view reframes the classic accuracy–robustness dilemma. The cost of higher accuracy is not an abstract fragility but an explicit redistribution of curvature: the network concentrates confidence by steepening loss gradients off the data manifold. Robustness falls because iterative attacks exploit exactly these steep off-manifold walls. SBDE, therefore, does more than reveal a trade-off—it localizes it. It shows that part of the accuracy–robustness tension is a question of \emph{where} the curvature lives and whether those directions can be controlled or projected away.

More broadly, this study suggests a methodological shift. Instead of seeking robustness only through adversarial training or regularization, we can design structured perturbations like SBDE that expose how geometry responds to optimization. Such probes transform adversarial attacks from threats into diagnostic instruments: they let us map the topology of the loss landscape and quantify how learning carves its valleys. In that light, the accuracy–robustness trade-off is not a paradox to be lamented but a geometric law to be understood and possibly engineered around.

In conclusion, SBDE and projection \(\Pi\) together form a practical geometry lab for modern neural networks. They allow us to observe, in controlled fashion, how sharpening accuracy reshapes curvature and how attacks exploit that structure. This approach offers a bridge between empirical robustness studies and theoretical geometry, grounding long-standing intuitions about the accuracy–robustness tension in directly measurable geometric evidence.


\bibliography{cas-refs}

@article{lecun2015deep,
  title={Deep learning},
  author={LeCun, Yann and Bengio, Yoshua and Hinton, Geoffrey},
  journal={Nature},
  volume={521},
  number={7553},
  pages={436--444},
  year={2015},
  url={https://webofscience.clarivate.cn/wos/alldb/full-record/WOS:000355286600030},
  doi={10.1038/nature14539}
}

@inproceedings{he2016deep,
  title={Deep residual learning for image recognition},
  author={He, Kaiming and Zhang, Xiangyu and Ren, Shaoqing and Sun, Jian},
  booktitle={Proc. IEEE Conf. Comput. Vis. Pattern Recognit.},
  pages={770--778},
  year={2016},
  url={https://webofscience.clarivate.cn/wos/alldb/full-record/WOS:000400012300083},
  doi={10.1109/CVPR.2016.90}
}

@article{dosovitskiy2021an,
  title={An image is worth 16x16 words: {Transformers} for image recognition at scale},
  author={Dosovitskiy, Alexey and Beyer, Lucas and Kolesnikov, Alexander and Weissenborn, Dirk and Zhai, Xiaohua and others},
  journal={arXiv preprint arXiv:2010.11929},
  year={2020}
}

@article{krizhevsky2017imagenet,
  title={{ImageNet} classification with deep convolutional neural networks},
  author={Krizhevsky, Alex and Sutskever, Ilya and Hinton, Geoffrey E},
  journal={Commun. ACM},
  volume={60},
  number={6},
  pages={84--90},
  year={2017},
  url={https://webofscience.clarivate.cn/wos/alldb/full-record/WOS:000402555400026},
  doi={10.1145/3065386}
}

@article{simonyan2015very,
  title={Very deep convolutional networks for large-scale image recognition},
  author={Simonyan, Karen and Zisserman, Andrew},
  journal={arXiv preprint arXiv:1409.1556},
  year={2014}
}

@article{szegedy2014intriguing,
  title={Intriguing properties of neural networks},
  author={Szegedy, Christian and Zaremba, Wojciech and Sutskever, Ilya and Bruna, Joan and Erhan, Dumitru and Goodfellow, Ian and Fergus, Rob},
  journal={arXiv preprint arXiv:1312.6199},
  year={2013}
}

@article{goodfellow2015explaining,
  title={Explaining and harnessing adversarial examples},
  author={Goodfellow, Ian J and Shlens, Jonathon and Szegedy, Christian},
  journal={arXiv preprint arXiv:1412.6572},
  year={2014}
}

@inproceedings{carlini2017towards,
  title={Towards evaluating the robustness of neural networks},
  author={Carlini, Nicholas and Wagner, David},
  booktitle={IEEE Symp. Secur. Priv. (SP)},
  pages={39--57},
  year={2017},
  url={https://webofscience.clarivate.cn/wos/alldb/full-record/WOS:000413081300003},
  doi={10.1109/SP.2017.49}
}

@inproceedings{papernot2016limitations,
  title={The limitations of deep learning in adversarial settings},
  author={Papernot, Nicolas and McDaniel, Patrick and Jha, Somesh and Fredrikson, Matt and Celik, Z Berkay and Swami, Ananthram},
  booktitle={IEEE Eur. Symp. Secur. Priv. (EuroS\&P)},
  pages={372--387},
  year={2016},
  url={https://webofscience.clarivate.cn/wos/alldb/full-record/WOS:000386286200024},
  doi={10.1109/EuroSP.2016.36}
}

@inproceedings{fawzi2018cvpr,
  title={Empirical study of the topology and geometry of decision boundaries},
  author={Fawzi, Alhussein and Moosavi-Dezfooli, Seyed-Mohsen and Frossard, Pascal},
  booktitle={Proc. IEEE Conf. Comput. Vis. Pattern Recognit. (CVPR)},
  pages={3762--3770},
  year={2018},
  url={https://openaccess.thecvf.com/content_cvpr_2018/papers/Fawzi_Empirical_Study_of_CVPR_2018_paper.pdf},
  doi={10.1109/CVPR.2018.00396}
}

@article{gilmer2018spheres,
  title={Adversarial Spheres},
  author={Gilmer, Justin and Ford, Luke and Carlini, Nicholas and Cubuk, Ekin D},
  journal={arXiv preprint arXiv:1801.02774},
  year={2018}
}

@inproceedings{stutz2019cvpr,
  title={Disentangling Adversarial Robustness and Generalization},
  author={Stutz, David and Hein, Matthias and Schiele, Bernt},
  booktitle={Proc. IEEE/CVF Conf. Comput. Vis. Pattern Recognit.},
  pages={6976--6987},
  year={2019},
  url={https://webofscience.clarivate.cn/wos/alldb/full-record/WOS:000542649300044},
  doi={10.1109/CVPR.2019.00714}
}

@article{akhtar2018threat,
  title={Threat of adversarial attacks on deep learning in computer vision: A survey},
  author={Akhtar, Naveed and Mian, Ajmal},
  journal={IEEE Access},
  volume={6},
  pages={14410--14430},
  year={2018},
  url={https://webofscience.clarivate.cn/wos/alldb/full-record/WOS:000429250000001},
  doi={10.1109/ACCESS.2018.2807385}
}

@inproceedings{ilyas2019adversarial,
  title={Adversarial examples are not bugs, they are features},
  author={Ilyas, Andrew and Santurkar, Shibani and Tsipras, Dimitris and Engstrom, Logan and Tran, Brandon and Madry, Aleksander},
  booktitle={Adv. Neural Inf. Process. Syst. (NeurIPS)},
  volume={32},
  year={2019},
  url={https://proceedings.neurips.cc/paper/2019/file/e2c420d928d4bf8ce0ff2ec19b371514-Paper.pdf}
}

@inproceedings{athalye2018obfuscated,
  title={Obfuscated gradients give a false sense of security: Circumventing defenses to adversarial examples},
  author={Athalye, Anish and Carlini, Nicholas and Wagner, David},
  booktitle={Proc. Int. Conf. Mach. Learn.},
  pages={274--283},
  year={2018},
  url={https://webofscience.clarivate.cn/wos/alldb/full-record/WOS:000683379200029}
}

@article{croce2020reliable,
  title={Reliable evaluation of adversarial robustness with AutoAttack},
  author={Croce, Francesco and Hein, Matthias},
  journal={Adv. Neural Inf. Process. Syst.},
  volume={33},
  pages={274--285},
  year={2020}
}

@article{tramer2020adaptive,
  title={On Adaptive Attacks to Adversarial Example Defenses},
  author={Tram{\`e}r, Florian and Carlini, Nicholas and Brendel, Wieland and Madry, Aleksander},
  journal={Adv. Neural Inf. Process. Syst.},
  volume={33},
  pages={1633--1645},
  year={2020},
  url={https://proceedings.neurips.cc/paper/2020/file/11f38f8ecd71867b42433548d1078e38-Paper.pdf}
}

@article{tsipras2019robustness,
  title={Robustness may be at odds with accuracy},
  author={Tsipras, Dimitris and Santurkar, Shibani and Engstrom, Logan and Turner, Alexander and Madry, Aleksander},
  journal={arXiv preprint arXiv:1805.12152},
  year={2018}
}

@inproceedings{zhang2019theoretically,
  title={Theoretically Principled Trade-off between Robustness and Accuracy},
  author={Zhang, Hongyang and Yu, Yaodong and Jiao, Jiantao and Xing, Eric P. and El Ghaoui, Laurent and Jordan, Michael I.},
  booktitle={Proc. Int. Conf. Mach. Learn. (ICML)},
  pages={7472--7482},
  year={2019},
  url={http://proceedings.mlr.press/v97/zhang19p/zhang19p.pdf}
}

@inproceedings{schmidt2018adversarially,
  title={Adversarially robust generalization requires more data},
  author={Schmidt, Ludwig and Santurkar, Shibani and Tsipras, Dimitris and Talwar, Kunal and Madry, Aleksander},
  booktitle={Adv. Neural Inf. Process. Syst.},
  volume={31},
  year={2018},
  url={https://proceedings.neurips.cc/paper/2018/file/f708f064faaf32a43e4d3c784e6af9ea-Paper.pdf}
}

@article{meng2022Adversarial,
  title={Adversarial robustness of deep neural networks: A survey from a formal verification perspective},
  author={Meng, Mark Huasong and Bai, Guangdong and Teo, Sin Gee and Hou, Zhe and Xiao, Yan and Lin, Yun},
  journal={IEEE Trans. Dependable Secure Comput.},
  year={2022},
  url={https://ieeexplore.ieee.org/document/9785704},
  doi={10.1109/TDSC.2022.3179131}
}

@article{wang2022Adversarial,
  title={Adversarial attacks and defenses in deep learning for image recognition: A survey},
  author={Wang, Jia and Wang, Chengyu and Lin, Qiuzhen and Luo, Chengwen and Luo, Chengwen and Li, Jianqiang},
  journal={Neurocomputing},
  volume={514},
  pages={162--181},
  year={2022},
  doi={10.1016/j.neucom.2022.09.004}
}

@article{madry2018towards,
  title={Towards Deep Learning Models Resistant to Adversarial Attacks},
  author={Madry, Aleksander and Makelov, Aleksandar and Schmidt, Ludwig and Tsipras, Dimitris and Vladu, Adrian},
  journal={arXiv preprint arXiv:1706.06083},
  year={2017}
}

@incollection{Kurakin2018Adversarial,
  title={Adversarial examples in the physical world},
  author={Kurakin, Alexey and Goodfellow, Ian J and Bengio, Samy},
  booktitle={Artificial intelligence safety and security},
  pages={99--112},
  year={2018},
  publisher={Chapman and Hall/CRC}
}

@article{Ising1925,
  title={Beitrag zur {Theorie} des {Ferromagnetismus}},
  author={Ising, Ernst},
  journal={Z. Phys.},
  volume={31},
  number={1},
  pages={253--258},
  year={1925},
  doi={10.1007/bf02980577}
}

@article{symmetry2025,
  title={Symmetry breaking in neural network optimization: insights from input dimension expansion},
  author={Zhang, Jun-Jie and Cheng, Nan and Li, Fu-Peng and Wang, Xiu-Cheng and Chen, Jian-nan and Pang, Long-Gang and Meng, Deyu},
  journal={npj Artif. Intell.},
  volume={1},
  number={1},
  pages={12},
  year={2025},
  url={https://www.nature.com/articles/s44387-025-00010-0.pdf},
  doi={10.1038/s44387-025-00010-0}
}

@article{zhang2024exploring,
  title={Exploring the uncertainty principle in neural networks through binary classification},
  author={Zhang, Jun-Jie and Chen, Jian-Nan and Meng, De-Yu and Wang, Xiu-Cheng},
  journal={Sci. Rep.},
  volume={14},
  number={1},
  pages={28402},
  year={2024},
  url={https://www.nature.com/articles/s41598-024-79028-4},
  doi={10.1038/s41598-024-79028-4}
}

@article{ZHANG2025112197,
  title={On the Uncertainty Principle of Neural Networks},
  author={Zhang, Jun-Jie and Zhang, Dong-Xiao and Chen, Jian-Nan and Pang, Long-Gang and Meng, Deyu},
  journal={iScience},
  volume={28},
  number={4},
  year={2025},
  url={https://www.cell.com/iscience/fulltext/S2589-0042(25)00458-4},
  doi={10.1016/j.isci.2025.112197}
}

@article{zhang2024nsr,
  title={Quantum-inspired analysis of neural network vulnerabilities: the role of conjugate variables in system attacks},
  author={Zhang, Jun-Jie and Meng, Deyu},
  journal={Natl. Sci. Rev.},
  volume={11},
  number={9},
  pages={nwae141},
  year={2024},
  url={https://academic.oup.com/nsr/article/11/9/nwae141/7644355?login=false},
  doi={10.1093/nsr/nwae141}
}

@article{Sun_2024,
  title={{JefiAtten}: an attention-based neural network model for solving {Maxwell}'s equations with charge and current sources},
  author={Sun, Ming-Yan and Xu, Peng and Zhang, Jun-Jie and Du, Tai-Jiao and Wang, Jian-Guo},
  journal={Mach. Learn.: Sci. Technol.},
  volume={5},
  number={3},
  pages={035055},
  year={2024},
  url={https://iopscience.iop.org/article/10.1088/2632-2153/ad6ee9/meta},
  doi={10.1088/2632-2153/ad6ee9}
}

@article{Zhang_2025,
  title={A Pedagogical Explanation of the Inherent Uncertainty Principle of Neural Networks from the Perspective of Gradient-Based Attacks},
  author={Zhang, Jun-Jie and Chen, Jian-Nan and Meng, Deyu},
  journal={Mod. Appl. Phys.},
  volume={16},
  number={01},
  pages={11304--11304},
  year={2025}
}

\end{document}